\pdfoutput=1
\documentclass[5p,twocolumn]{elsarticle}
\usepackage[utf8]{inputenc}
\usepackage{lineno} 
\usepackage{times}
\usepackage{epsfig}
\usepackage{graphicx}
\usepackage{amsmath}
\usepackage{amssymb}
\usepackage{booktabs}
\usepackage[utf8]{inputenc}
\usepackage{url}
\usepackage{color}
\usepackage{multirow}

\usepackage{mathrsfs}
\usepackage{dsfont}
\DeclareMathAlphabet{\mathpzc}{OT1}{pzc}{m}{it}
\newcommand{\Indicator}{{\mathds{I}}}

\newcommand{\bP}{\mbox{\boldmath$P$}}
\newcommand{\bp}{\mbox{\boldmath$p$}}
\newcommand{\bq}{\mbox{\boldmath$q$}}
\newcommand{\bn}{\mbox{\boldmath$n$}}

\newcommand{\bz}{\mbox{\boldmath$z$}}

\modulolinenumbers[5]

\journal{Computer Vision and Image Understanding}

\begin{document}

\begin{frontmatter}

\title{Hand Pose Estimation through Semi-Supervised and Weakly-Supervised Learning }

\author[ulyon]{Natalia Neverova\corref{mycorrespondingauthor}}
\cortext[mycorrespondingauthor]{Corresponding author (currently at Facebook AI Research, Paris)}
\ead{neverova@fb.com}
\author[ulyon]{Christian Wolf}
\author[awabot]{Florian Nebout}
\author[uog]{Graham W.~Taylor}

\address[ulyon]{Université de Lyon, INSA-Lyon, CNRS, LIRIS, F-69621, France}
\address[awabot]{Awabot SAS, France}
\address[uog]{School of Engineering, University of Guelph, Canada\par\noindent\hspace*{10pt}}



\begin{abstract}
  We propose a method for hand pose estimation based on a deep regressor trained on two different kinds of input. Raw depth data is fused with an intermediate representation in the form of a segmentation of the hand into parts. This intermediate representation contains important topological information and provides useful cues for reasoning about joint locations. The mapping from raw depth to segmentation maps is learned in a semi- and weakly-supervised way from two different datasets: (i) a synthetic dataset created through a rendering pipeline including densely labeled ground truth (pixelwise segmentations); and (ii) a dataset with real images for which ground truth joint positions are available, but not dense segmentations. Loss for training on real images is generated from a patch-wise restoration process, which aligns tentative segmentation maps with a large dictionary of synthetic poses. The underlying premise is that the domain shift between synthetic and real data is smaller in the intermediate representation, where labels carry geometric and topological meaning, than in the raw input domain. Experiments on the NYU dataset \cite{TompsonSIGGRAPH2014} show that the proposed training method decreases error on joints over direct regression of joints from depth data by 15.7\%.
\end{abstract}


\end{frontmatter}


\section{Introduction}

\begin{figure}[t] \centering
\includegraphics[width=7.8cm]{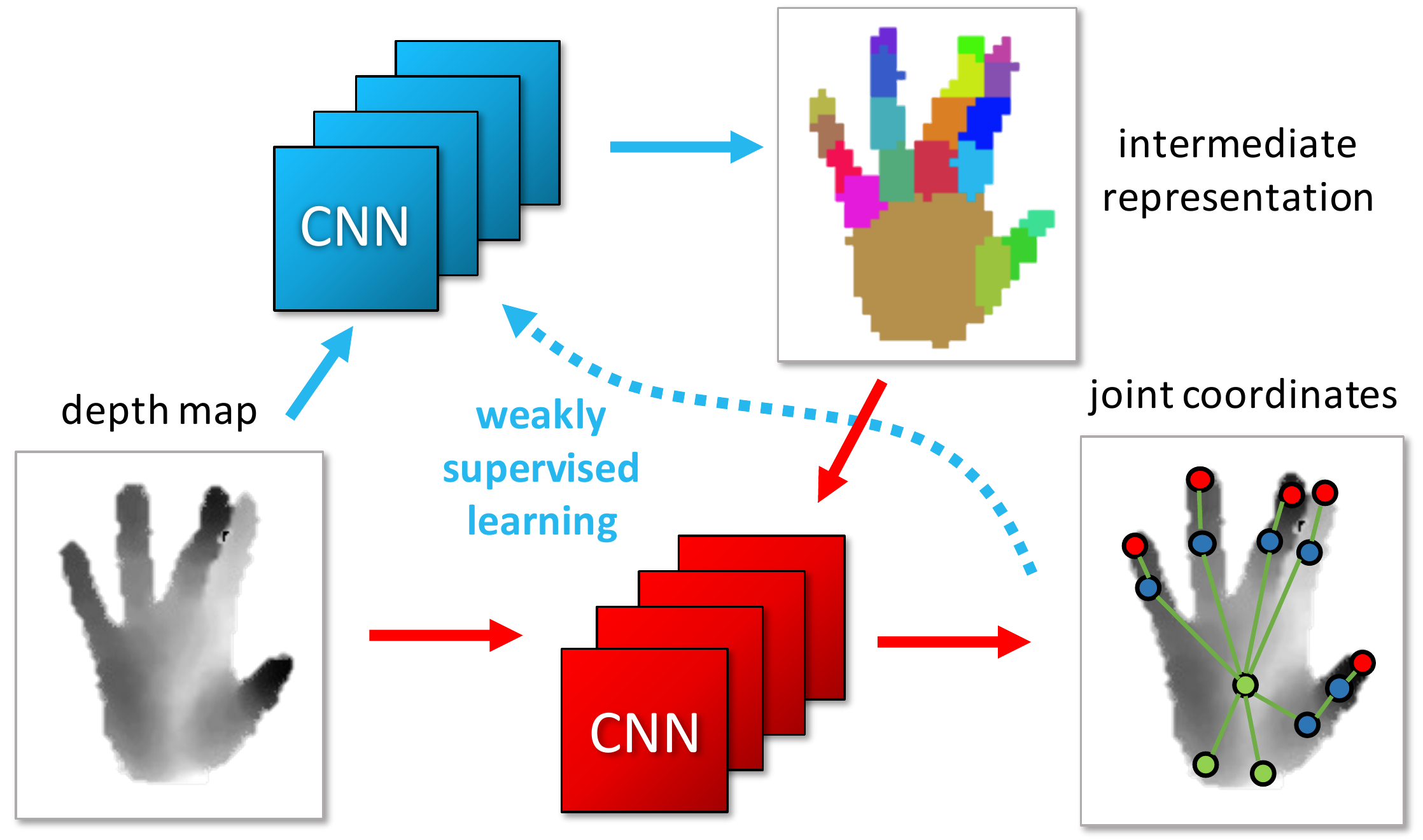}
\caption{A rich intermediate representation is fused with raw input for hand joint regression. The intermediate representation is learned in a semi/weakly-supervised setting from real and synthetic data (see Figure \ref{fig:scheme} for an illustration of the training procedure). The input image is from the NYU dataset \cite{TompsonSIGGRAPH2014}.\label{fig:teaser}}\vspace*{-5pt}
\end{figure}

\noindent
Hand pose estimation and tracking from depth images, i.e.~the estimation of joint positions of a human hand, is a first step for various applications: hand gesture recognition, human-computer interfaces (moving cursors and scrolling documents), human-object interaction in virtual reality settings and many more. While real-time estimation of full-body pose is now available in commercial products, at least in cooperative environments \cite{Shotton2011}, the estimation of hand pose is more complex. In situations where the user is not directly placed in front of the computer, and therefore not close to the camera, the problem is inherently difficult. In this case, typically the hand occupies only a small portion of the image, and fingers and finger parts are only vaguely discernible.

Existing discriminative solutions to articulated pose from depth maps (see Section \ref{sec:relatedwork} for a full description of related work) have concentrated on two different strategies: (i) early work first contructed an intermediate representation of the body or hand into parts, from which joints were then estimated in a second step \cite{Shotton2011,KeskinAkarun2012}; (ii) subsequent work proceeded by direct regression from depth either through heat maps or to joint coordinates \cite{TompsonSIGGRAPH2014,TompsonJainNIPS2014,ChenYuille2014,ShottonCVPR2012,ChenYuille2015,chu2016,Tang2015,
oberweger2015feedback}.

There is clear interest in the second strategy, i.e.~direct regression of joint positions, with a large body of recent work pointing in this direction. One of the reasons is that intermediate representations are only really efficient when training is restricted to synthetic data, where intermediate labels can be easily obtained. On real data, precise and dense annotations like part segmentations are difficult to come by, with the exception of finger painting datasets \cite{Sharp2015}. On the other hand, the amount of information passed to the training algorithm is significantly higher in the case of the intermediate segmentation: several bits \emph{per pixel} of the input image vs.~2 or 3 real values \emph{per joint and per image}. In this paper we argue that this advantage is important, and we propose a new model for regression as well as a semi-supervised and weakly-supervised training algorithm which allows us to extract this information automatically from real data.

The novelty of our approach compared to existing work lies in the integration of the intermediate representation as a \emph{latent variable} during training: no dense annotation on real data is required, only joint positions are needed.


In our target configuration, pose estimation is performed frame-by-frame without any dynamic information. A model is learned from two training sets: (i) a (possibly small) set of real images acquired with a consumer depth sensor. Ground truth joint positions are assumed to be available, for instance obtained by multi-camera motion capture systems. (ii) a second (and possibly very large) set of synthetic training images produced from 3D models by a rendering pipeline, accompanied by dense ground truth in the form of a segmentation into hand parts. This ground truth is easy to come by, as it is usually automatically created by the same rendering pipeline.
The main arguments we develop are the following:\vspace*{-7pt}
\begin{itemize}
\item an intermediate representation defined as a segmentation into parts contains rich structural and topological information, since the label space itself is structured. Labels have adjacency, topological and geometric relationships, which can be leveraged and translated into loss for weakly-supervised training;\vspace*{-7pt}
\item a regression of joint positions is easier and more robust if the depth input is combined with a rich semantic representation like a segmentation into parts, provided that this semantic segmentation is of high fidelity (see Figure \ref{fig:teaser}).\vspace*{-7pt}
\end{itemize}
We show that the additional information passed to the training algorithm is able to improve pose regression performance. A key component to obtaining this improvement is obtaining reliable segmentations for real data.  While purely supervised training on synthetic data has proven to work well for full-body pose estimation \cite{Shotton2011,KeskinAkarun2012}, hand pose estimation is known to require real data captured from depth sensors for training \cite{TangYuKim2013,TangChangTejaniKim2014,TompsonSIGGRAPH2014} due to low input resolution and data quality. Manually annotating dense segmentations of large datasets is not an option, and estimating segmentation maps from ground truth joint positions is unreliable in the case of low quality images. We propose a semi/weakly supervised setting in order to tackle this problem, where this intermediate representation is learned from densely labeled synthetic depth images as well as from real depth images associated with ground truth joint positions.

In particular, the proposed training method exploits the rich geometrical and topological information of the intermediate representation. During the training process, predicted segmented patches from real images are aligned with a very large dictionary of labelled patches extracted from rendered synthetic data. The novelty here lies in the fact that we do not match input patches but patches taken from the intermediate representation, which include the to-be-inferred label and its local context.

We call the proposed training method \emph{weakly supervised}; since part of the ground truth data only contains joint locations and not the dense labels of part segmentations used during inference. The method is also \emph{semi-supervised}, as part of the dataset is fully labeled, while another part is not.

The paper is organized as follows.
Section \ref{sec:relatedwork} discusses related work.
Section \ref{sec:ourmeth} introduces the model and the semi/weakly supervised learning setting.
Section \ref{sec:architectures} gives details about the deep architectures employed in the experiments.
Section \ref{sec:experiments} explains the experimental setup and provides results.
Section \ref{sec:conclusion} concludes.
     
\section{Related work}
\label{sec:relatedwork}

\noindent
Compared to the study of hand pose, a much larger body of work has focused on full-body pose estimation. We draw influence from this literature and therefore include it in our brief review.\smallskip

\noindent
\textbf{Learning ---}
The majority of recent work on pose estimation is based on machine learning. Body part segmentation as an intermediate representation for joint estimation from depth images was successfully used by Shotton et al.~\cite{Shotton2011}, where random forests were trained to perform pixel-wise classification. This was adapted to hand pose estimation by Keskin et al.~\cite{KeskinAkarun2012}, where an additional pose clustering step was introduced. Later, Sun et al.~\cite{Sun2015} proposed a cascaded regression framework where positions of hand joints are predicted in a hierarchical manner adapted to the kinematic structure of a hand.
In work by Tang et al.~\cite{TangYuKim2013}, a random forest performs different tasks at different levels: viewpoint clustering in higher levels, part segmentation in intermediate levels, and joint regression in lower levels. An explicit transfer function is learned between synthetic and real data. In follow-up work by Tang et al.~\cite{TangChangTejaniKim2014}, a latent regression forest is learned, which automatically extracts a hierarchical and topological model of a human hand from data. Instead of pixel-wise voting, the forest is descended a single time starting from the center of mass of the hand. The traditional split nodes in the RF are accompanied by division nodes, which trigger parallel descents of sub-trees for multiple entities (joints or groups of joints).  Li et al.~\cite{li2015} went further in this direction and proposed a method where the topological model of a hand is learned jointly with hand pose estimation.
Later work uses deep networks for body part segmentation from RGB images \cite{VarolRomeroSchmid2017} or depth images \cite{JiuWolfTaylorBaskurt2013}.

Recent work estimates joint positions by regression with deep convolutional neural nets~\cite{TompsonSIGGRAPH2014,ChenYuille2014,TompsonJainNIPS2014,toshev2014deeppose,JainTompsonTaylor2013,oberweger2015feedback,Ge2017,Sinha2017}.
Typically such models have been trained to produce heatmaps encoding the joint positions as spatial Gaussians, though direct regression to an encoding of joints has also been attempted.
Recently, training with a combination of classification and regression losses for producing heatmaps has been proven particularly effective \cite{BulatTzimiropoulosECCV2016}.

Post-processing to enforce structural constraints is based on graphical models~\cite{JainTompsonTaylor2013,TompsonJainNIPS2014,ChenYuille2014,ChenYuille2015} or inverse kinematics~\cite{TompsonSIGGRAPH2014}.
In a work by Oberweger et al.~\cite{DeepPrior} a deep learning framework is regularized by a bottleneck layer, forcing the network to model underlying structure of joint positions. Ge at al.~\cite{ge2016} proposed a multi-view deep learning framework based on feature extraction from several projections of a point cloud representing a depth image of a hand.
Stacked hourglass networks \cite{NewellYDHourglass2016} introduce a series of modules, each of which perform convolutions followed by deconvolutions with skip connections. The different modules allow the method to model context, in the spirit of auto-context models for semantic segmentation \cite{Pinheiro14}.

In a recent overview~\cite{Supancic2015}, deep learning models were shown to perform the best among existing approaches, but still far from human performance. In this context of data-driven methods, there were a number of recent works dedicated to automating data labeling \cite{oberweger2016data} and weakly-supervised learning for sparsely annotated videos~\cite{koller2016}.\smallskip

\noindent
\textbf{Graphical models ---}
In recent work by Chen and Yuille~\cite{ChenYuille2014}, a graphical model is implemented with deep convolutional nets, which jointly estimate unary terms, given evidence of joint types and positions, and binary terms, modelling relationships between joints. This work was later extended in~\cite{ChenYuille2015} for modelling occluding parts by introducing an additional connectivity prior.
Tompson et al.~\cite{TompsonJainNIPS2014} jointly learn a deep full-body part detector with a Markov Random Field which models spatial priors. Joint learning is achieved by designing the priors as convolutions and implementing inference as forward propagation approximating a single step in a message passing algorithm.
Alternatively, Chu et al.~\cite{chu2016} proposed a method for learning structured representations by capturing dependencies between body joints during training with introduced geometrical transform kernels.\smallskip

\footnotetext{\url{https:/www.youtube.com/watch?v=7GMiExWKM8c}}

\noindent
\textbf{Top down methods ---}
A different group of methods is based on top-down processes which fit 3D models to image data. A method by Oikonomidis et al.~\cite{Oikonomidis2011} is based on pixelwise comparison of rendered and observed depth maps. Inverse rendering of a generative model including shading and texture is used for model fitting in~\cite{DeLaGorceFleetParagios2011}. Several works~\cite{Liang2013,tagliasacchi2015} employ ICP for hand pose reconstruction and 3D fingertip localization under spatial and temporal constraints.
More recent work by Tang et al.~\cite{Tang2015} is based on hierarchical sampling optimization, where a sequence of predictors is aligned with kinematic structure of a hand.

A hybrid method by Qian et al.~\cite{qian2014} for real-time hand tracking uses a simple hand model consisting of a number of spheres and combines a gradient-based discriminative step with stochastic optimization.
More recent work by Sharp et al.~\cite{Sharp2015} directly exploits a hand mesh consisting of triangles and vertices to formulate a multi-level discriminative strategy followed by generative model-based refinement.
In~\cite{SridharOulasvirta2014}, a person-specific and hybrid generative/discriminative model is built for a system using five RGB cameras plus a ToF camera.
In the spirit of hybrid methods, Oberweger et al.~\cite{oberweger2015feedback} proposed a neural model including a feedback loop based on iterative image generation from obtained predictions followed by self-correction.\smallskip

Top down down methods can be very accurate, if the geometric models are flexible enough and can closely fit the real geometric data. In recent work, geometric models are adapted to each instance (body or hand). In \cite{ShottonSiggraph2016}, for instance, the two fitting steps (pose estimation and shape estimation) are integrated into a single joint optimization procedure. The downside of these models is their requirement for initialization. Hybrid models combine generative and discriminative steps, for instance by performing discriminative initialization followed by fitting of a generative model.

Our work can be seen as an improvement of the discriminative stage which could be augmented with a generative counterpart of choice.

\noindent
\textbf{Correspondence ---} Pose estimation has been attempted by solving correspondence problems in other work. In a method by Taylor et al.~\cite{TaylorShottonFitzgibbon2012}, correspondence between depth pixels and vertices of an articulated 3D model is learned. In~\cite{Athitsos2003}, approximate directed Chamfer distances are used to align observed edge images with a synthetic dataset.\smallskip

Recently, model fitting and training a network on dense correspondances between the input and the template has  proven to be extremely efficient in the context of facial landmark localization \cite{densereg}. Along these lines,
creation of large-scale synthetic datasets for human pose estimation \cite{VarolRomeroSchmid2017} will facilitate adaptation of dense prediction methods to the human pose. Our work is essentially motivated by the same idea of exploiting dense signals as a source of universal and rich supervision.


\noindent
\textbf{Other work ---}
Work on applications other than pose estimation share similarities with our method. Patchwise alignment of segmentations in a transductive learning setting has been performed on 3D meshes~\cite{XuToG2014}, matching real unlabelled shape segments to shapes from a large labelled database. A large number of candidate segments is created and the optimal segmentation is calculated solving an integer problem. Our method can be viewed as a kind of multi-task learning, a topic actively pursued by the deep vision community~\cite{devries2014multi-task, zhang2014facial}. While these techniques rely on subnetworks that share parameters, our approach does not use weight sharing, instead, it co-ordinates subnetworks via a patchwise restoration process and a joint error function.

Our work bears a certain resemblance to the recently proposed $N^4$-Fields~\cite{GaninLempitskyACCV2014}. This method, which was proposed for different applications, also combines deep networks and a patchwise nearest neighbor (NN) search. However, whereas in~\cite{GaninLempitskyACCV2014}, NN-search is performed in a feature space learned by deep networks, our method performs NN-search in a patch space corresponding to semantic segmentation learned by deep networks. This part of our work also bears some similarity to the way label information is integrated in structured prediction forests~\cite{kontschieder2011}, although no patch alignment with a dictionary is carried out there.
       
\section{Semi-supervised and weakly-supervised learning of joint regression}
\label{sec:ourmeth}

\begin{figure}[t] \centering
\includegraphics[height=4cm]{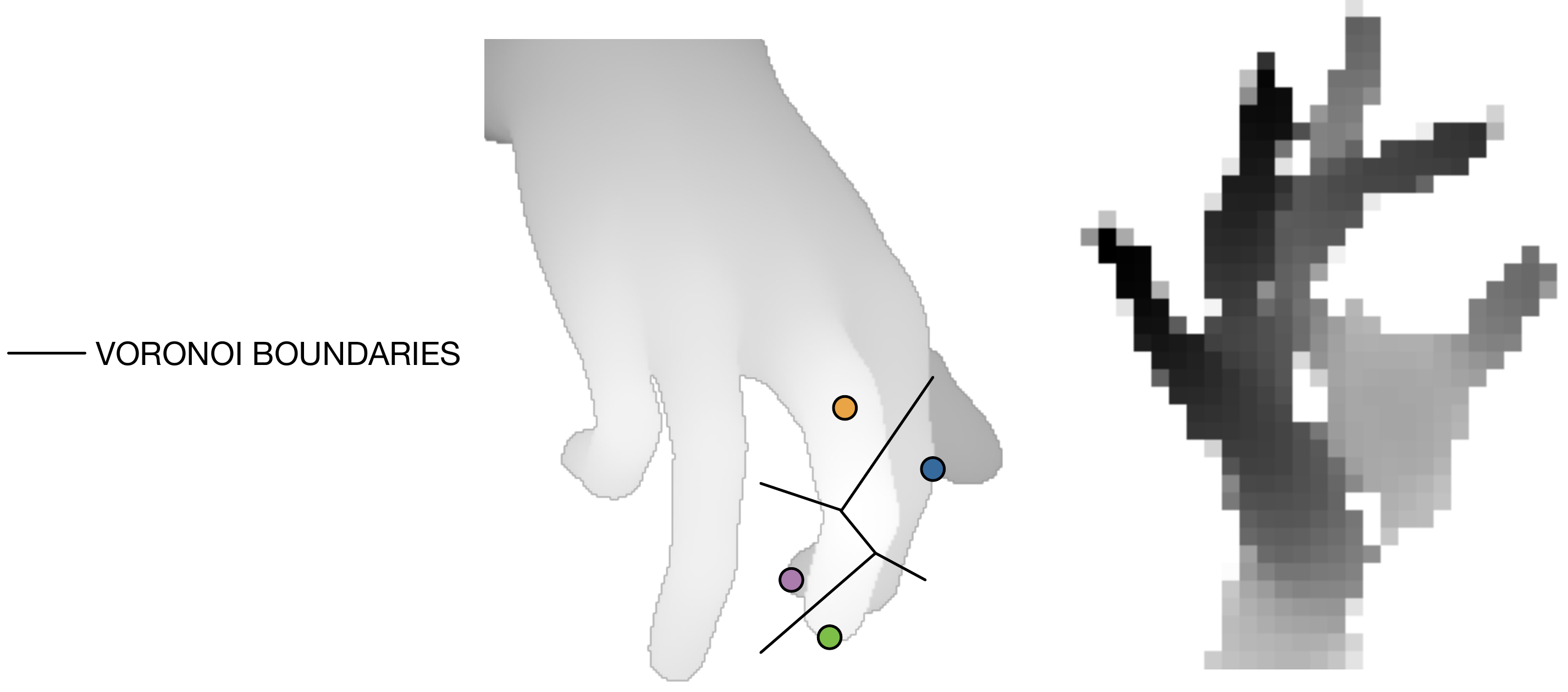}\\
\caption{\label{fig:compat_represention}
Complementarity of segmentation maps and key points: the space occupied by the index finger and the ring finger is difficult to segment, as seen by the Voronoi borders (on the left); depth values are often very similar and close for different fingers at realistic sensor resolutions (taken from the NYU Hand Pose dataset, on the right).}
\end{figure}

\begin{figure*}[t] \centering
\includegraphics[width=17.5cm]{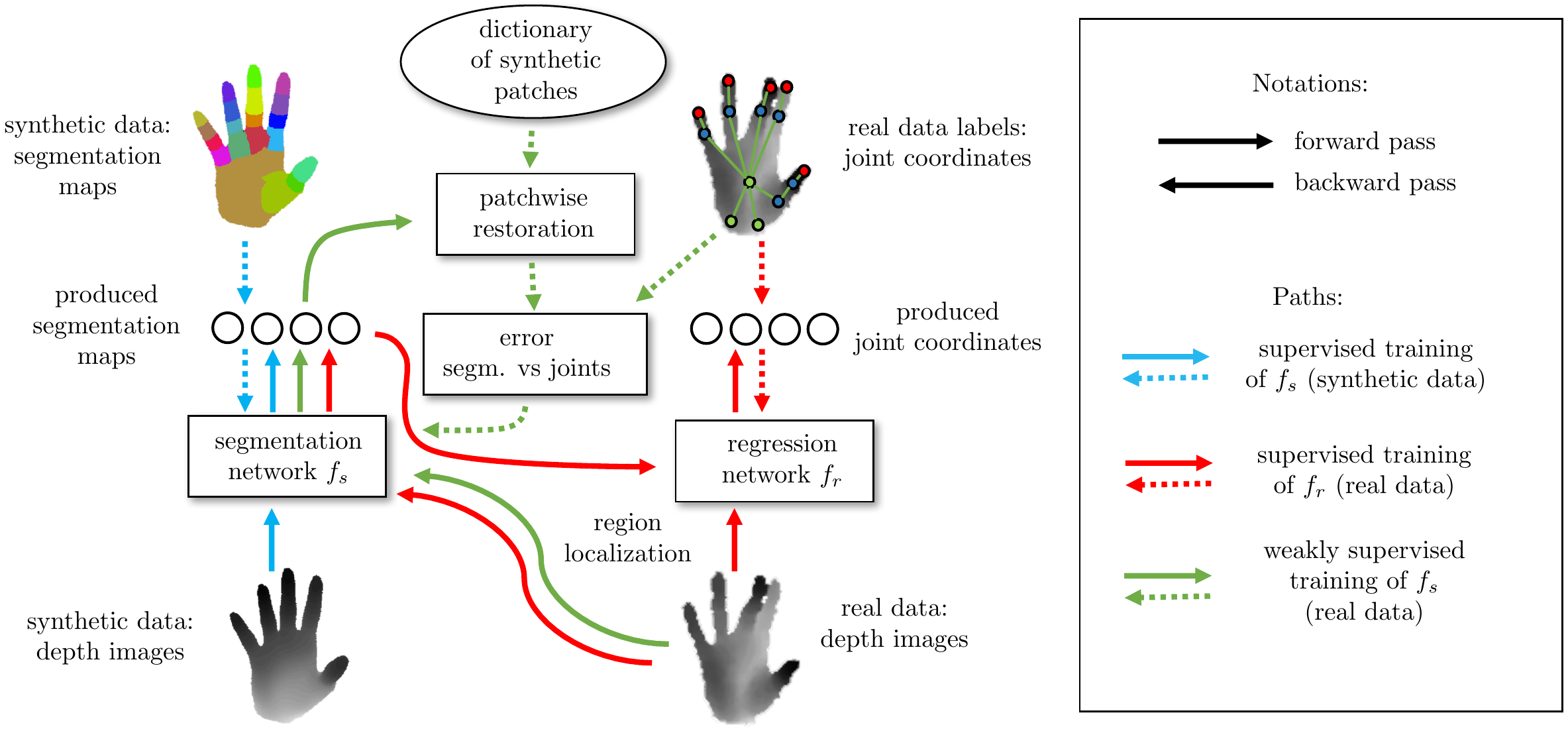}
\caption{A functional overview of the method. Blue data flow corresponds to supervised training of the segmentation network $f_s$. Green flow corresponds to weakly supervised training of $f_s$. Red arrows show supervised training of the regressor $f_r$ (for more detail see a supplementary video available online \protect\footnotemark).}
\label{fig:scheme}
\end{figure*}

\noindent
When prediction models are learned automatically, representation learning and full end-to-end training are often desirable and described as the holy grail in machine learning. This approach indeed benefits from properties, such as freedom of the training procedure to explore and find the best representation and a lower burden on the scientist or practitioner, who is not required to handcraft representations from knowledge of the application and/or the input data domain.

In practice, however, depending on the difficulty of the problem, the depth of the model and the amount of available training data, it might be suboptimal to give the model complete freedom over the intermediate representations it explores. Problems can either arise from overfitting or from suboptimal solutions found by the minimization procedure. Instead of falling back to handcrafting feature representations, solutions can be intermediate supervision, or a decomposition approach, where an intermediate representation is temporarily imposed during a pre-training step before full end-to-end training.

In this work we propose an intermediate representation in the form of a segmentation map. In contrast to traditional decomposition approaches, the intermediate representation is available as additional information to the final regressor, which also receives raw input.
Our intermediate representation is a segmentation into 20 parts, illustrated in Figure \ref{fig:teaser}. 19 parts correspond to finger parts, one part to the palm (see Figure \ref{fig:histsegments} for exact definitions). The background is considered to be subtracted in a preprocessing step and is not part of the segmentation process.

This dense segmentation is complementary to groundtruth key points and the former is very difficult to obtain from the latter in the case of strong auto occlusions. This is illustrated in Figure \ref{fig:compat_represention}a: the space occupied by the index finger and the ring finger is difficult to segment. Resorting to spatial distances alone, traditionally done in easier settings, fails due to complex curved shapes of fingers and the low amount of keypoints which can be obtained with automatic methods (2 points per finger in the NYU dataset \cite{TompsonSIGGRAPH2014}). In this context, a Voronoi diagram gives regions which are very different from the complex finger regions. Depth values might theoretically help, but in practice this fails since the values are often very similar and close for different fingers at realistic sensor resolutions, as shown in Figure \ref{fig:compat_represention}b taken from the NYU dataset.

Compared to the initial input depth image, this representation has several important advantages:
\begin{itemize}\vspace*{-5pt}
\item The part label space is characterized by strong topological properties. In contrast to other semantic segmentation problems (for instance, semantic full scene labeling), strong neighborhood relationships can be defined on the label space. They can be leveraged to restore noisy part label images and to generate a loss function for training. This property serves as a key component of creating reliable estimators and for performing transfer from synthetic to real images.\vspace*{-5pt}
\item For a given view and pose, the part label itself carries strong geometrical information: the label alone of a given pixel is a very strong prior on the position of the pixel in 3d. This provides important cues for regression to the desired joint positions. This property will also be exploited in our system to motivate patchwise searches in label space.
\end{itemize}

\noindent
In our setting, the intermediate representation is available during training time for the synthetic data only. For real data and during test time, it is automatically inferred. More precisely, our training dataset is organized into two partitions: a set of real depth images with associated ground truth joint positions, and a second set of synthetic depth images with associated ground truth label images (segmentation maps in the intermediate representation). We will denote by $D^{(j)}$ the $j^{th}$ pixel in image~$D$.

The synthetic images have been rendered from different 3D hand models using a rendering pipeline (details will be given in Section \ref{sec:experiments}). As in \cite{SridharOulasvirta2014}, we also sample different pose parameters and hand shape parameters. Variations in viewpoints and hand poses are obtained taking into account physical and physiological constraints.  In contrast to other weakly-supervised or semi-supervised methods, for instance \cite{TangYuKim2013}, we do not suppose that any ground truth data for the intermediate representation is available for the real training images. Manually labelling segmentations is extremely difficult and time consuming. Labelling a sufficiently large number of images is hardly practical.

We do, however, rely on ground truth for joint positions, which can be obtained in several ways: In \cite{Shotton2011}, external motion capture using markers is employed. In \cite{TompsonSIGGRAPH2014}, training data is acquired from multiple views from three different depth sensors and an articulated model is fitted offline.

The functional decomposition of the method as well as the dataflow during training and testing are outlined in Figure~\ref{fig:scheme}.
The goal is to regress joint positions from input depth maps, and to this end, the proposed method leverages two different mappings learned on two training sets. A \textit{segmentation network} learns a mapping $f_s(\cdot,\theta_s)$ from raw depth data to intermediate segmentation maps, parametrized by a parameter vector $\theta_s$. A \textit{regression network} learns a mapping $f_r(\cdot,\theta_r)$ from raw depth data combined with segmentation maps to joint positions, parametrized by a vector $\theta_r$.

The training procedure uses both synthetic and real data, and proceeds in three steps:
\begin{enumerate}
\item First, the segmentation network $f_s$ is pre-trained on synthetic training data in a supervised way using dense ground truth segmentations, resulting in a prediction model $f_s(\cdot,\theta_s)$. The parameters are learned minimizing classical negative log-likelihood (NLL). This training step is shown as blue data flow in Figure~\ref{fig:scheme}.
\item Then the prediction model for $f_s$ is fine-tuned in a subsequent step by weakly supervised training on real data, resulting in a refined prediction model $f_s(\cdot,\theta'_s)$. This step, shown as green data flow in Figure~\ref{fig:scheme}, is described in more detail in Subsection \ref{sec:weaklysupervised}.
\item Finally, the regression network $f_r$ is trained on real data. It is implemented as a mapping $f_r(\cdot,\theta_r)\colon (D,N) \to \bz$ from a full size input depth image $D$ and a full size segmentation map $N$ to a joint location vector $\bz$. Parameters $\theta_r$ are trained classically by minimizing the $L_2$ norm between output joint positions and ground truth joint positions. This training step is shown as red data flow in Figure~\ref{fig:scheme}.
\end{enumerate}
Subsection \ref{sec:weaklysupervised} provides more details on step 2, the weakly supervised training procedure. The actual deep architectures employed will be described separately in Section \ref{sec:architectures}.



\begin{table}[!t] \centering
\vspace*{-2mm}
\begin{tabular}{c|c|c}
\toprule
Layer & Filter size / units & Pooling \\
\noalign{\smallskip}
\hline
\multicolumn{3}{c}{Segmentation network $f_s$} \\
\hline
Depth input & $48{\times}48$ &  - \\
Conv.~layer 1 & $32{\times}5{\times}5$ & $2{\times}2$\\
Conv.~layer 2 & $64{\times}5{\times}5$ &  $2{\times}2$\\
Conv.~layer 3 & $128{\times}5{\times}5$ &  $1{\times}1$\\
Hidden layer & $500$                            & -\\
Hidden layer & $500$                            & -\\
Output          & $20{\times}48{\times}48$    & - \\
\hline
\multicolumn{3}{c}{Regression network $f_r$}\\
\hline
Depth input     & $24{\times}24$& - \\
Conv.~layer c1 & $32{\times}3{\times}3$ & $2{\times}2$\\
Conv.~layer c211, c221 & $16{\times}1{\times}1$&\\
Conv.~layer c212, c222 & $8{\times}3{\times}3$ &\\
Pooling p231 &-&$2{\times}2$\\
Conv.~layer 232 &$8{\times}1{\times}1$ &\\
Conv.~layer 241 &$16{\times}1{\times}1$&\\
Hidden layer fc & $1200$             & - \\
Output             & $14{\times}3$ & - \\
 \hline
\end{tabular}
\caption{\label{table:networkparameters} Hyper-parameters chosen for the deep networks.}
\end{table}

\begin{figure*}[t!] \centering
\includegraphics[width=15cm]{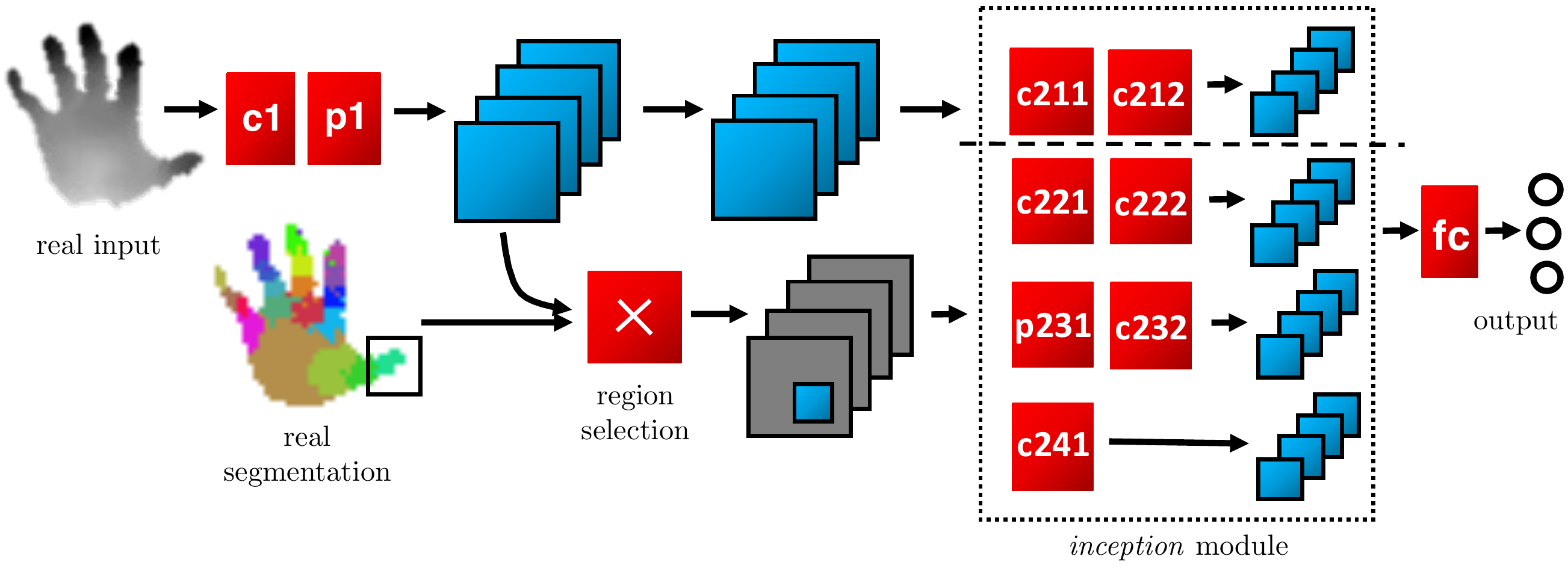}
\caption{\label{fig:regressor}Organization of the regression network $f_r$. All functional modules are shown in red and the produced feature maps are shown in blue. The grey areas correspond to masked regions on the feature maps. Masked areas are shown as rectangles in the figure, but are of general shape.}
\end{figure*}

\subsection{Weakly supervised fine-tuning of the segmentation network}
\label{sec:weaklysupervised}

\noindent
Supervised pre-training of the segmentation network results in a prediction model $f_s(\cdot,\theta_s)$. To address the domain shift between synthetic data and real data shot with depth sensors, the model is fine-tuned by training on real data. Since no ground truth segmentation maps exist for this data, we generate a loss function for training based on two sources:
\begin{itemize}
\item sparse information in the form of ground truth joint positions, and
\item  \textit{a priori} information on the local distribution of part labels on human hands through a patch-wise restoration process, which aligns noisy predictions with a large dictionary of synthetic poses.
\end{itemize}
The weakly supervised training procedure is shown as green data flow in Figure~\ref{fig:scheme}. Each real depth image is passed through the pre-trained segmentation network $f_s$, resulting in a segmentation map. This noisy predicted map is restored through a restoration process $f_{nn}$ described further below. The quality of the restored segmentation map is estimated by comparing it to the set of ground truth joint positions for that image. In particular, for each joint, a corresponding part-label is identified, and the barycenter of the corresponding pixels in the segmentation map is calculated. A rough quality measure for a segmentation map can be given as the sum (over joints) of the $L_2$ distances between barycenters and ground truth joint positions. The quality measure is used to determine whether the restoration process has lead to an improvement in segmentation quality, i.e.~whether the barycenters of the restored map are closer to the ground truth joint positions than the barycenters of the original prediction. For images where this is the case, the segmentation network is updated for each pixel, minimizing NLL using the labels of the restored map as artificial ``ground truth''.

\begin{figure*}[t!] \centering
\includegraphics[width=0.85\linewidth]{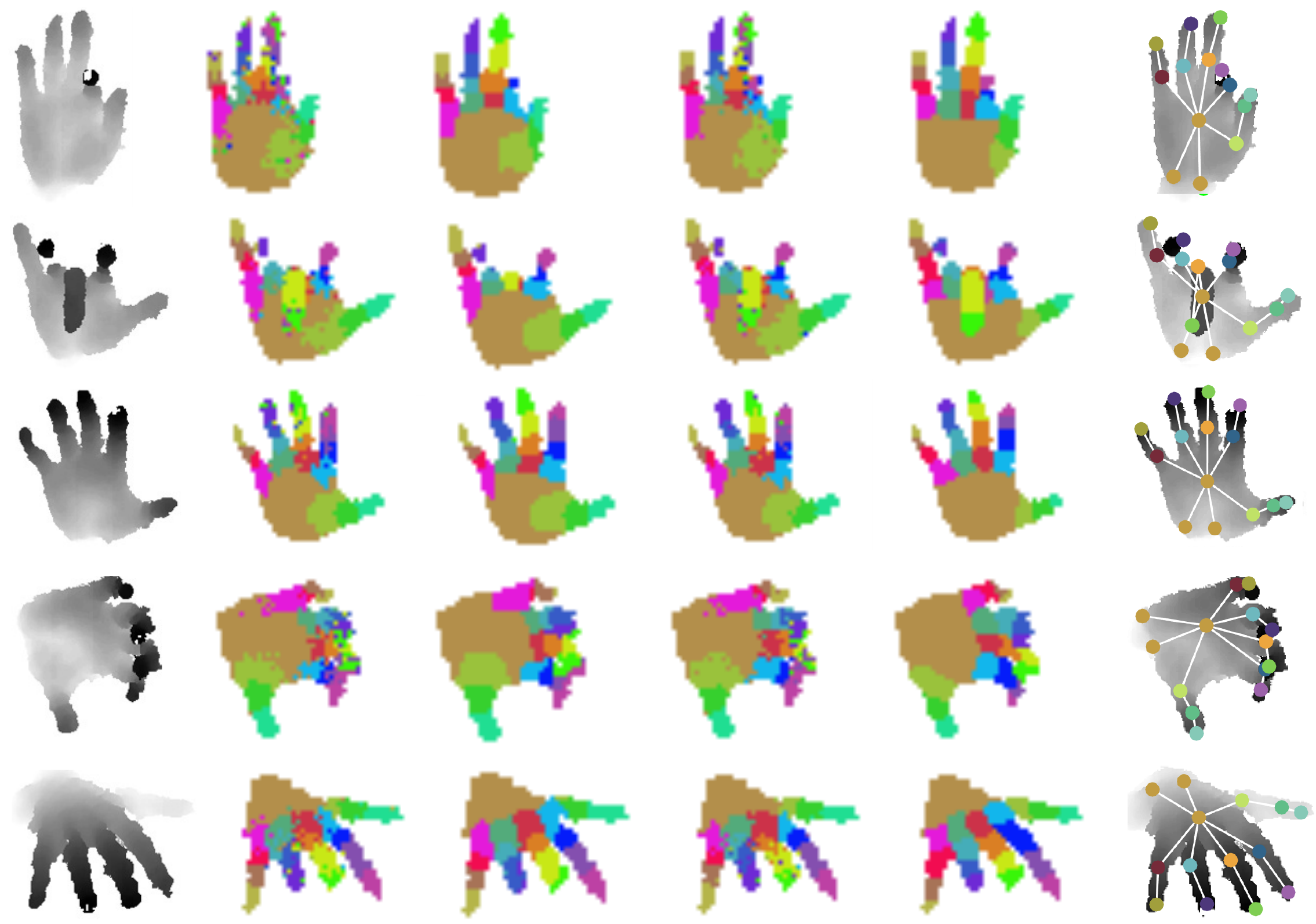}
\\
(a) \hspace{2.0cm}
(b) \hspace{2.0cm}
(c) \hspace{2.1cm}
(d) \hspace{2.0cm}
(e) \hspace{2.2cm}
(f) \hspace{2.0cm}
\\
\caption{Different segmentation results: (a) input image; (b) output of the segmentation network after supervised training; (c) after restoration; (d) output of the segmentation network after joint training; (e) ground truth segmentation; (f) estimated joint positions. The image itself was \textit{not} part of the training set.
\label{fig:restexamples}}
\end{figure*}

\subsection{Patchwise restoration}
\label{sec:restoration}

\noindent
We proceed patchwise, extracting patches of size $P{\times}P$ from a large set of synthetic segmentation images, resulting in a dictionary of patches $\bP{=}\{\bp^{(l)}\}, \ l{\in}\{1\dots N\}$. In our experiments, we used patches of size $27{\times}27$ and a dictionary of $36$ million patches is extracted from the training set (see Section \ref{sec:experiments}). As also reported in \cite{SridharOulasvirta2014}, the range of poses which can occur in natural motion is extremely large, making full global matching with pose datasets difficult. This motivates our patch-based approach, which aims to match at a patch-local level rather than matching whole images.

A given real input depth image $D$ is aligned with this dictionary in a patchwise process using the intermediate representation. For each pixel $j$, a patch $\bq^{(j)}$ is extracted from the segmentation produced by the network $f_s$, and the nearest neighbor $\mathbf{\nu}(\bq^{(j)})$ is found by searching the dictionary $\bP$:
\begin{equation} \bn^{(j)} \triangleq \mathbf{\nu}(\bq^{(j)}) = \arg \min_{\bp^{(l)} \in \bP} d_H (\bq^{(j)},\bp^{(l)}) ,\
\label{eq:flann}
\end{equation} where $d_H(\bq,\bp)$ is the Hamming distance between the two patches $\bq$ and $\bp$. The search performed in (\ref{eq:flann}) can be calculated efficiently using KD-trees.

In a na\"ive setting, a restored label for pixel $j$ could be obtained by choosing the label of the center of the retrieved patch $\bn^{(j)}$. This, however, leads to noisy restorations, which suggest the need for spatial context. Instead of chosing the center label of each patch only, we propose to use all of the labels in each patch. For each input pixel, the nearest neighbor results in a local window of size $W{\times}W$ are integrated. In particular, for a given pixel $j$, the assigned patches $\bn^{(k)}$ of all neighbors $k$ are examined and the position of pixel $j$ in each patch is calculated. A label is estimated by voting, resulting in a mapping $f_{nn}(.)$:
\begin{equation}
f_{nn}(\bq^{(j)}) =
\arg \max_l \sum_{k\in W{\times}W}
\Indicator(l=\bn^{(k,j@k)})
\label{sec:localintegration}
\end{equation}
where
$\bn^{(k)}{=}\mathbf{\nu}(\bq^{(j)})$ is the nearest neighbor result for pixel $k$,
$\bn^{(j,m)}$ denotes pixel $m$ of patch $\bn^{(j)}$,
$\Indicator({\omega}){=}1$ if $\omega$ holds and $0$ else,
and the expression $j@k$ denotes the position of pixel $j$ in the patch centered on neighbor $k$.

This integration bears some similarity to \cite{GaninLempitskyACCV2014}, where information of nearest neighbor searches is integrated over a local window, albeit through averaging a continuous mapping. It is also similar to the patchwise integration performed in structured prediction forests \cite{kontschieder2011}.

If real-time performance is not required, the patch alignment process in Equation \ref{eq:flann} can be regularized with a graphical model including pairwise terms favoring consistent assignments, for instance, a Potts model or a term favoring patch assignments with consistent overlaps. Interestingly, this produces only very small gains in performance, especially given the higher computational complexity. Moreover, the gains vanish if local integration (Equation \ref{sec:localintegration}) is added.
More information is given in Section \ref{sec:experiments}.



\section{Architectures}
\label{sec:architectures}

\noindent
The structure of the segmentation network $f_s$ is motivated by the idea of performing efficient pixelwise image segmentation preserving the original resolution of the input, and is inspired by \emph{OverFeat} networks which were proposed for object detection and localization \cite{OverFeat,NeverovaWolfTaylorNeboutACCV2014}. The network consists of 3 convolutional layers,
followed by a fully connected layer (see Table~\ref{table:networkparameters}). Max pooling, which typically follows convolutional layers, results in downsampling of feature maps, destroying precise spatial information. Instead, we perform pooling over $2{\times}2$ overlapping regions produced by shifting the feature maps obtained at the previous step by a single pixel along one or another axis. As opposed to patchwise training of pixel classification based on its local neighborhood, such an architecture is more computationally efficient, as it benefits from dense computations at earlier layers. Compared to recently introduced fully-convolutional \cite{long_shelhamer_fcn2015} and deconvolutional networks \cite{noh2015learning}, which tackle a similar problem in the context of semantic segmentation, the proposed network requires less upsampling and interpolation.

The regression network, taking a depth image as a single input and producing 3 coordinates for a given joint, also incorporates the information provided by the segmentation network during training. Structurally, it resembles an Inception module \cite{GoogLeNet,NetworkInNetwork} where the output of the first convolutional layer after max pooling as passed through several parallel feature extractors capturing information at different levels of localization. The organization of this module is shown in Figure \ref{fig:regressor}.
and the corresponding parameters are, as before, provided in Table \ref{table:networkparameters}.
The output of the first convolutional layer c1 (followed by pooling p1) is aligned with the segmentation maps produced by the segmentation network. From each feature map, for a given joint we extract a localized region of interest filtered by the mask of a hand part to which it belongs (or a number of hand parts which are naturally closest to this joint). These masks are calculated by performing morphological opening on the regions having the corresponding class label in the segmentation map.
Once the local region is selected, the rest of the feature map area is set to 0. The result, along with the original feature maps is then fed to the next layer, i.e.~an Inception module.
The rest of the training process is organized in such a way that the network's capacity is split between global structure of the whole image and the local neighborhood, and a subset of Inception $3{\times}3$ filters is learned specifically from the local area surrounding the point of interest.

Experiments showed that individual networks for each joint do perform better than networks sharing parameters over joints. We conjecture, that sharing parameters would be the optimal choice for smaller amounts of training data. As the number of examples increases, separating networks allows all layers to pick up the fine nuances required for regressing each individual joint.

Both networks have ReLU activation functions at each layer and employ batch normalization \cite{BatchNormalization}. The regression network during test time uses the batch normalization parameters estimated on the training data, however, in the segmentation network, the batch normalization is performed across all pixels from the same image, for both training and test samples.

\section{Experimental results}
\label{sec:experiments}

\begin{table*} \centering
\begin{tabular}{lrrrr}
\toprule
Method & \multicolumn{2}{c}{--- per pixel---} & \multicolumn{2}{c}{--- per class ---}
\\
\midrule
No restoration
& 51.03 & & 39.38
\\
NN-search --- no integration
& 48.76 && 39.72
\\
NN-search --- integration w. equation (\ref{sec:localintegration})
& \textbf{54.55} & \textbf{(+3.52)} & \textbf{46.38} & \textbf{(+7.00)}
\\
CRF -- Potts-like model
& 53.10 & (+2.07) & 43.64 & (+4.26)
\\
CRF -- Hamming distance on overlapping patch area
& 52.45 & (+1.42) & 42.68 & (+3.30)
\\
\bottomrule
\end{tabular}
\caption{Restoration (=segmentation) accuracy on 100 manually labelled images of the NYU dataset (=NYU-100). \label{table:restperformance}}%
\end{table*}

\begin{figure*}[ht] \centering
\begin{tabular}{cc}
    \begin{minipage}{12cm}
    \includegraphics[width=12cm]{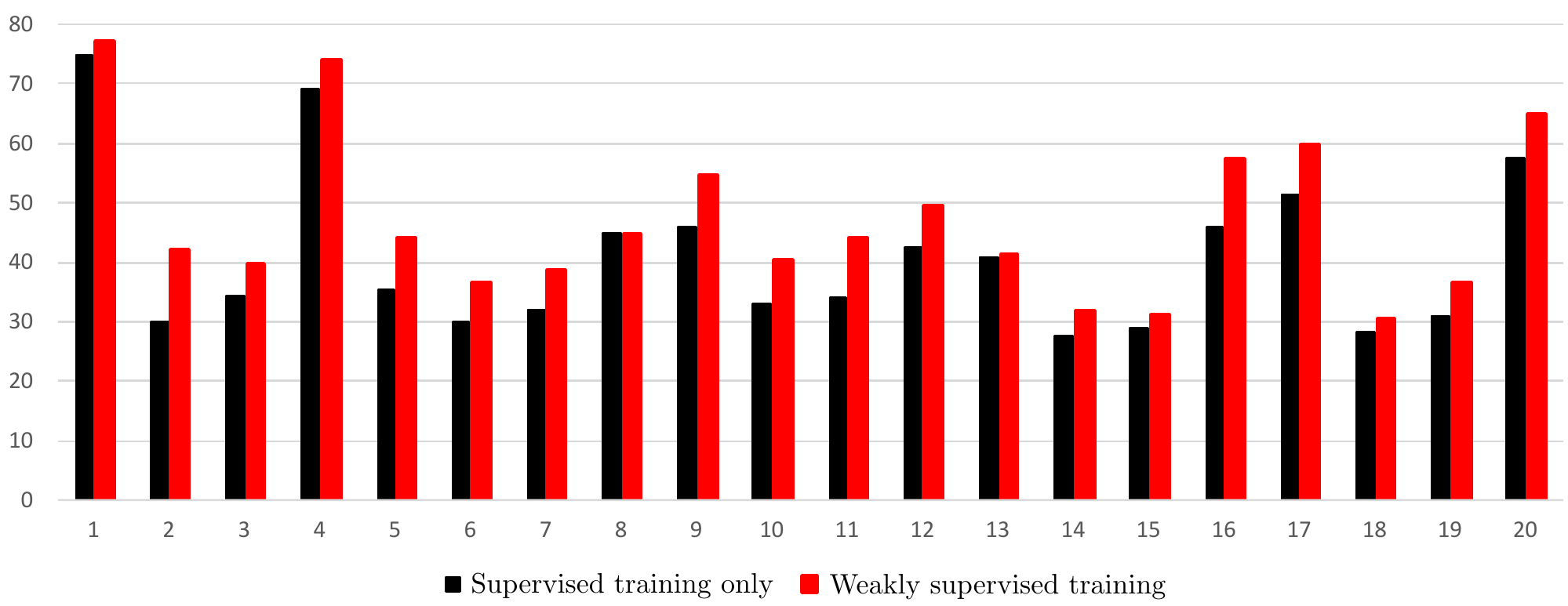}
    \end{minipage}\hspace*{5mm}
    &
    \begin{minipage}{3cm}
    $\phantom{t}$\vspace*{-1cm}\\
    \includegraphics[width=3cm]{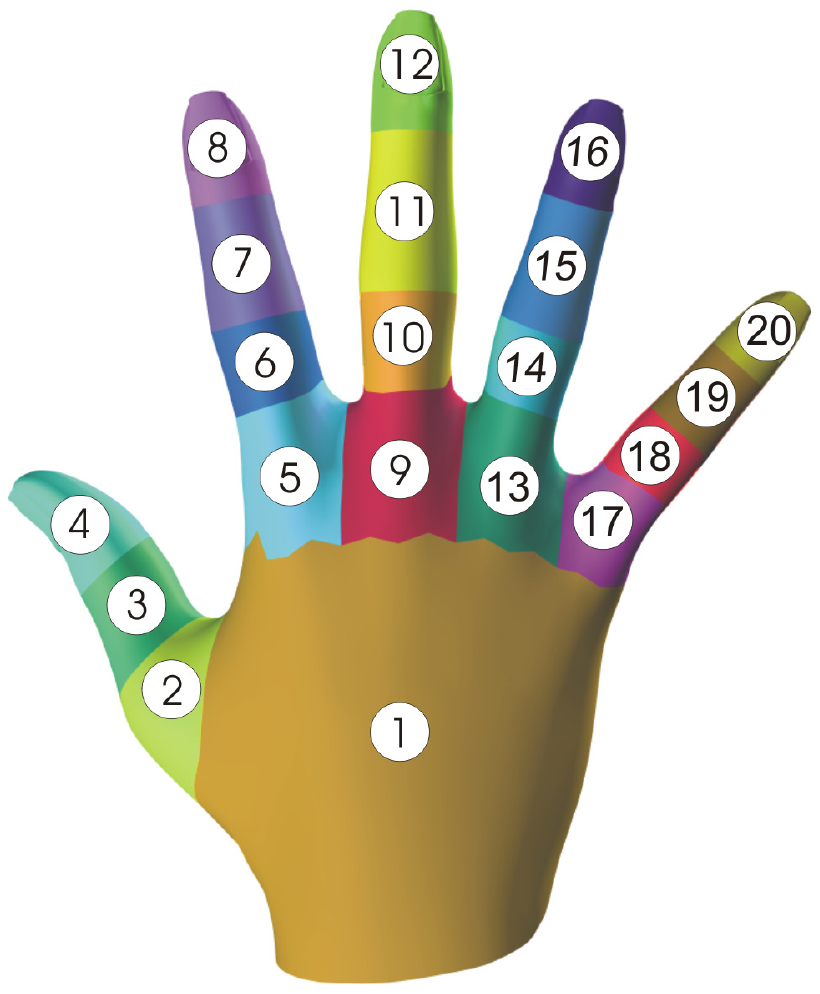}
    \end{minipage}
    \\
\end{tabular}
\caption{\label{fig:histsegments} Classification accuracy of the segmentation network $f_s$ for supervised training only (blue), and for weakly supervised training (red), where $1\ldots20$ is the number of a segment shown in the hand legend at the right.}
\end{figure*}

\noindent
We evaluated the proposed method on the \textit{NYU Hand Pose Dataset}, which was published in \cite{TompsonSIGGRAPH2014} and is publicly available\footnote{\url{http://cims.nyu.edu/~tompson/NYU_Hand_Pose_}\\\url{Dataset.htm}}. It comprises 70,000 images captured with a depth sensor in VGA resolution accompanied by ground truth annotations of positions of hand joints.



A video illustrating the results of our method is available online\footnote{\url{https://www.youtube.com/watch?v=7GMiExWKM8c}}. It shows the obtained pose as well as the intermediate representation during testing and during training, i.e.~after restoration.

To train the segmentation network, the synthetic training images were selected from our own dataset consisting of two subsets: (i) 170,974 synthetic training images rendered using the commercial software ``Poser'', including ground truth; (ii) 500 labelled synthetic images plus ground truth reserved for testing.

In our experiments, we extract hand images of normalized metric size (taking into account depth information) and resize them to $48{\times}48$ pixels. The data is preprocessed by local contrast normalization with a kernel size of 9. To be able to predict absolute values for z-coordinates, the subtracted depth is then added back to the network output.
For supervised training of the segmentation network and the regression network, the full set of 170,974 training images is used. A third of this set is used to extract patches for the patch alignment mapping $f_{nn}$, giving a dictionary of 36M patches of size $27{\times}27$ extracted from 56,991 images. Local integration as given in equation (\ref{sec:localintegration}) was done using windows of size $W{\times}W = 17{\times}17$.

The segmentation network was initially trained for 100 epochs with SGD, batch size 1, initial learning rate 0.1 and learning rate decay $10^{-5}$ on the synthetic data and then finetuned for additional 10 epochs on a mixture of synthetic and restored real segmentations with the ratio 9:1.

Finally, the regression network is trained by gradient descent using the Adam \cite{Adam} update rule, with learning rate set 0.05 and batch size of 64.

ConvNets were implemented using the Torch7 library \cite{CollobertTorch2011}. NN-search using KD-trees  was performed using the FLANN library \cite{MujaLoweFLANN2009}.
\medskip

\begin{figure*}[h!] \centering
\includegraphics[width=\linewidth]{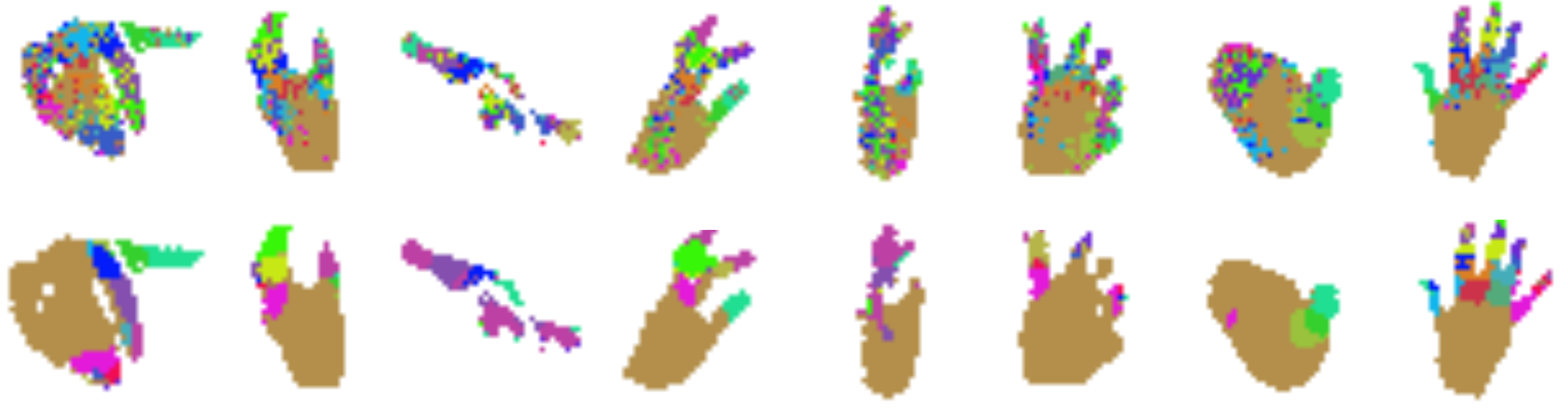}\\
(a) \hspace{1.75cm}
(b) \hspace{1.75cm}
(c) \hspace{1.75cm}
(d) \hspace{1.75cm}
(e) \hspace{1.75cm}
(f) \hspace{1.75cm}
(g) \hspace{1.75cm}
(h)
\caption{Visualization of the automatic rejections made by the quality check described in section \ref{sec:weaklysupervised}. Examples of raw noisy segmentations of real images are shown on top and the corresponding maps after restoration are shown at bottom. These restoration failure cases may happen due to artifacts in depth maps (c), (e), (f), increased distance between the camera and the hand (g) and particularly noisy initial predictions (a), (d), (h). As a result, the restoration process may result in lower recall of finger segments  (a)-(g) or, less often, false detections (h). Both cases can be automatically detected.
\label{fig:segfault}}
\end{figure*}

\begin{table*} \centering
\begin{tabular}{lcrrrr}
\toprule
Method & Datasets used & \multicolumn{2}{c}{--- per pixel---} & \multicolumn{2}{c}{--- per class ---}
\\
\midrule
Fully supervised training only & synth.~segmentations
& 51.03 && 39.38
\\
Semi-/weakly-supervised training & synth.~segmentations + real joint positions
& \textbf{57.18}  & (+6.15)& \textbf{47.20} & (+7.82)
\\
\bottomrule
\end{tabular}
\caption{The contribution of semi-/weakly-supervised training on segmentation accuracy. \label{table:weaklysupervisedperformance}}%
\end{table*}

\subsection{Segmentation performance}

\noindent
To evaluate the performance of the various segmentation methods, we manually labelled 100 images from the NYU dataset and report accuracy per pixel and per class in Table \ref{table:restperformance}. These 100 images were solely used for evaluation and never entered any training procedure. Recall that pure segmentation performance is of course not the goal of this work (or of the targeted application), it is given as additional information only.

Purely supervised training on the synthetic dataset gives poor segmentation performance of $51.03\%$ accuracy per pixel. We emphasize once more that training was performed on synthetic images while we test on real images, thus of an unseen distribution. This domain shift is clearly a problem, as accuracy on the synthetic dataset is very high, $90.16\%$. Using patchwise restoration of the predicted real patches with a large dictionary of synthetic patches gives a performance increase of $+3.5$ percentage points per pixel and $+7$ percentage points per class. This corroborates our intuition that the intermediate representation carries important structural information. Integration of patch-labels over a local window with equation (\ref{sec:localintegration}) is essential-- pure NN-search without integration performs poorly.

Figure \ref{fig:restexamples} provides examples of input depth maps and different segmentations: after synthetic pre-training only, after off-line restoration and a prediction after weakly-supervised finetuning. We can see that in a majority of cases the restoration output is very close to ground truth maps. Figure \ref{fig:histsegments} shows the distribution of the error over the different hand parts. The improvement is consistent, and high on the important fingertips.

Restoration failure cases, corresponding to 5-10\% of the real data, are demonstrated and explained in Figure \ref{fig:segfault}. At the fine-tuning stage, these samples are filtered out and excluded from training (see Section 3.1 for more detail).

We also compared local integration of equation (\ref{sec:localintegration}) to potentially more powerful regularization methods by implementing a CRF-like discrete energy function. The goal of the optimization problem is to regularize the patchwise restoration process described in Section \ref{sec:restoration}. Instead of chosing the nearest neighbor in patch space for each pixel as described in equation (\ref{sec:localintegration}), a solution is searched which satisfies certain coherence conditions over spatial neighborhoods. To this end, we create a global energy function $E(x)$ defined on a 2D-lattice corresponding to the input image to restore:
$$
E(x) = \sum_i u(x_i) + \alpha \sum_{i\sim j} b(x_i,x_j)
$$
where $i{\sim}j$ indices neighbors $i$ and $j$. Each pixel $i$ is assigned a discrete variable $x_i$ taking values between $1$ and $N{=}10$, where $x_i{=}l$ signifies that for pixel $i$ the $l{-}th$ nearest neighbor in patch space is chosen. For each pixel $i$, a nearest neighbor search is performed using KD-trees and a ranked list of $N$ neighbors is kept defining the label space for this pixel. The variable $N$ controls the degree of approximation of the model, where $N{=}\infty$ allows each pixel to be assigned every possible patch of the synthetic dictionary.

The unary data term $u(x)$ guides the solution towards approximations with low error. It is defined as the Hamming distance between the original patch and the synthethic patch.

\begin{table*}[t!] \centering
\begin{tabular}{l|c|c}
\toprule
\hspace*{33mm}Method & 2D error, mm & 3D error, mm\\
\midrule
Tompson et al. \cite{TompsonSIGGRAPH2014} & $7.1$ & $28.8$  \\
Oberweger et al. (DeepPrior) \cite{DeepPrior}		& $14.8$  &  $19.8$  \\
Oberweger et al. \cite{oberweger2015feedback} & $12.4$  &  $16.0$  \\
Bouchacourt et al. (DISCO) \cite{Bouchacourt2016} & -- & $20.7$  \\
Xu et al. (Lie-X) \cite{Xu2016}		     & -- &  $14.5$  \\
Zhou et al. (DeepModel) \cite{Zhou2016}      & -- &  $16.9$  \\
Deng et al. (Hand3D) \cite{Deng2017} & -- &  $17.6$  \\
Guo et al. (REN) \cite{Guo2017}          & -- & $13.4$  \\
Wan et al. (Crossing Nets) \cite{Wan2017}   & -- &  $15.5$  \\
Fourure et al. (JTSC) \cite{Fourure2017}            & $8.0$ &  $16.8$  \\
Zhang et al. \cite{Zhang2017}			     & -- &  $18.3$  \\
Madadi et al. \cite{Madadi2017}		     & -- &  $15.6$  \\
Oberweger et al. (DeepPrior++) \cite{Oberweger2017}     & -- &  $12.3$  \\
\midrule
Our baseline: direct regression (a) 		         & $13.3$  & $17.3$ \\
Our baseline: cascade direct regression (two steps) (b)           & $12.6$   & $16.9$ \\
Our baseline: semi-supervised network (c)	& $12.1$ &  $16.5$ \\
Our method: semi/weakly supervised network (d)	& $11.2$ &  $14.8$ \\
\bottomrule
\end{tabular}
\caption{Joint position estimation error on the NYU Hand Pose dataset (some values were estimated from plots if authors did not provide numerical values). Baseline (a) corresponds to the case of a single regression network where the segmentation results are not used. The cascaded baseline (b) is similar in architecture but features an additional refinement step (as proposed in \cite{DeepPrior}). Method (c) is based on the full pipeline with no restoration employed during training, i.e.~the segmentation network is trained on synthetic data in supervised way and not fine-tuned on the real samples. Finally, method (d) corresponds to our proposed solution.
\label{fig:jointposperformance}}
\end{table*}

We tested two different pairwise terms $b(x_i,x_j)$:

\begin{description}
\item[Potts-like terms] --- a Potts model classically favors equality of labels of neighboring sites. In our setting, we favor equality of the center pixels of the two patches assigned to $x_i$ and $x_j$.

\item[Patch-overlap distance] --- the alternative pairwise term is defined as the Hamming distance between the two synthethic patches defined by $x_i$ and $x_j$, in particular, the distance restricted to the overlapping area.
\end{description}
Inference was performed through message passing using the open-GM library \cite{AndresBeierKappesOpenGM2012}, and the hyper-parameter $\alpha$ was optimized on a hold-out set. Interestingly, local patch integration with equation (2) outperformed the combinatorial models significantly while at the same time being much faster. We conjecture that the reason is that our patchwise integration corresponds to a high-order potential (\ref{sec:localintegration}), whereas the binary terms in the CRF model are poorer. Calculating the global solution of poorer model seems to be less efficient than locally solving a high order problem. We see this as a further indication of the strong topological information carried by the label space of the intermediate representation.

Table \ref{table:weaklysupervisedperformance} shows the contribution of semi/weakly-supervised training, where sparse annotation (joint positions) are integrated into the training process of the segmentation network $f_s$. This procedure achieves an improvement of $+6.15$ percentage points per pixel and $+7.82$ percentage points per class, an essential step in learning an efficient intermediate representation.

At first sight it might seem to be odd that the pretrained predictor (+6.15pp w.r.t.t. baseline) is better than the off-line restoration process (+3.52pp), on whose results it has been trained. However, let us recall that the predictor also uses a second source of information, namely the joint locations available on real images during the weakly-supervised fine-tuning step. Rejecting bad segmentations during training is responsible for the difference. \medskip

\begin{figure*}[t!] \centering
\includegraphics[height=0.4\linewidth]{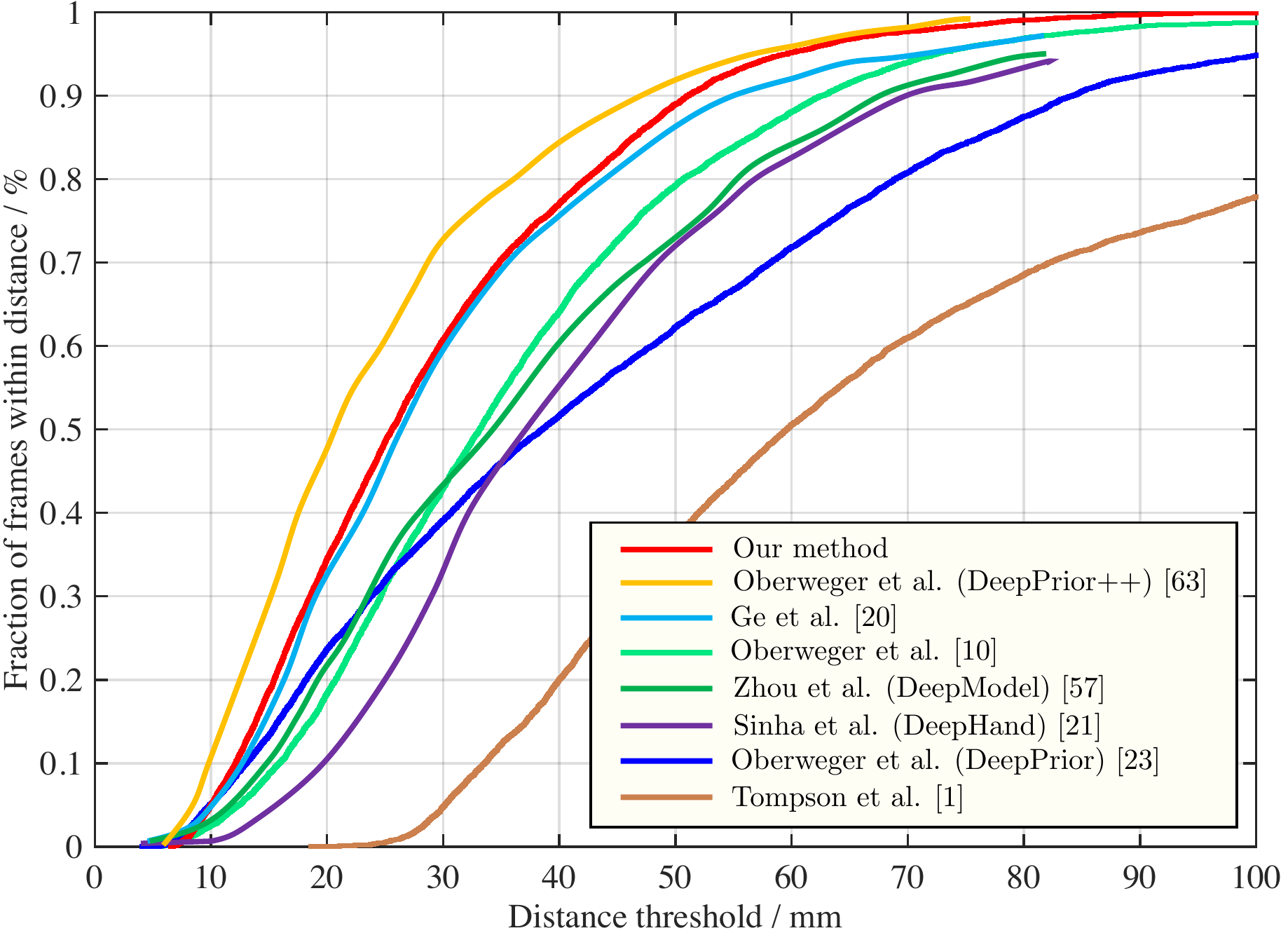}\\
\caption{\label{fig:plotmean}Joint estimation accuracy for the 3D case expressed as propotion of frames where all joints are localized within a given distance threshold in mm.}
\end{figure*}

\subsection{Hand joint position estimation}

\noindent
Table \ref{fig:jointposperformance} illustrates the effect of incorporating segmentation information on the performance of the regression network, i.e. on the error in joint positions --- the main goal of this work. In the bottom part of the table we compare the proposed method to our own baselines, which are based on an ablation study of the proposed network. In particular, we compare with direct regression without the intermediate representation, and with a version where the intermediate representation is learned in a supervised way only.

The first part of the table provides comparison with the literature and is based on publicly available data (predictions of uv-coordinates of joint locations) released by the authors of corresponding methods.
The error is  expressed as mean distance in mm (in 2D or 3D) between predicted position of each joint and its ground truth location. In comparison to a single network regressor, the 2D mean error was improved by $15.7\%$. The second best model not involving segmentation (cascade regression) was inspired by the work of \cite{toshev2014deeppose} and more recent \cite{DeepPrior}, where the initial rough estimation of hand joints positions is then improved by zooming in.

\begin{figure*}[p!] \centering
\includegraphics[width=\linewidth]{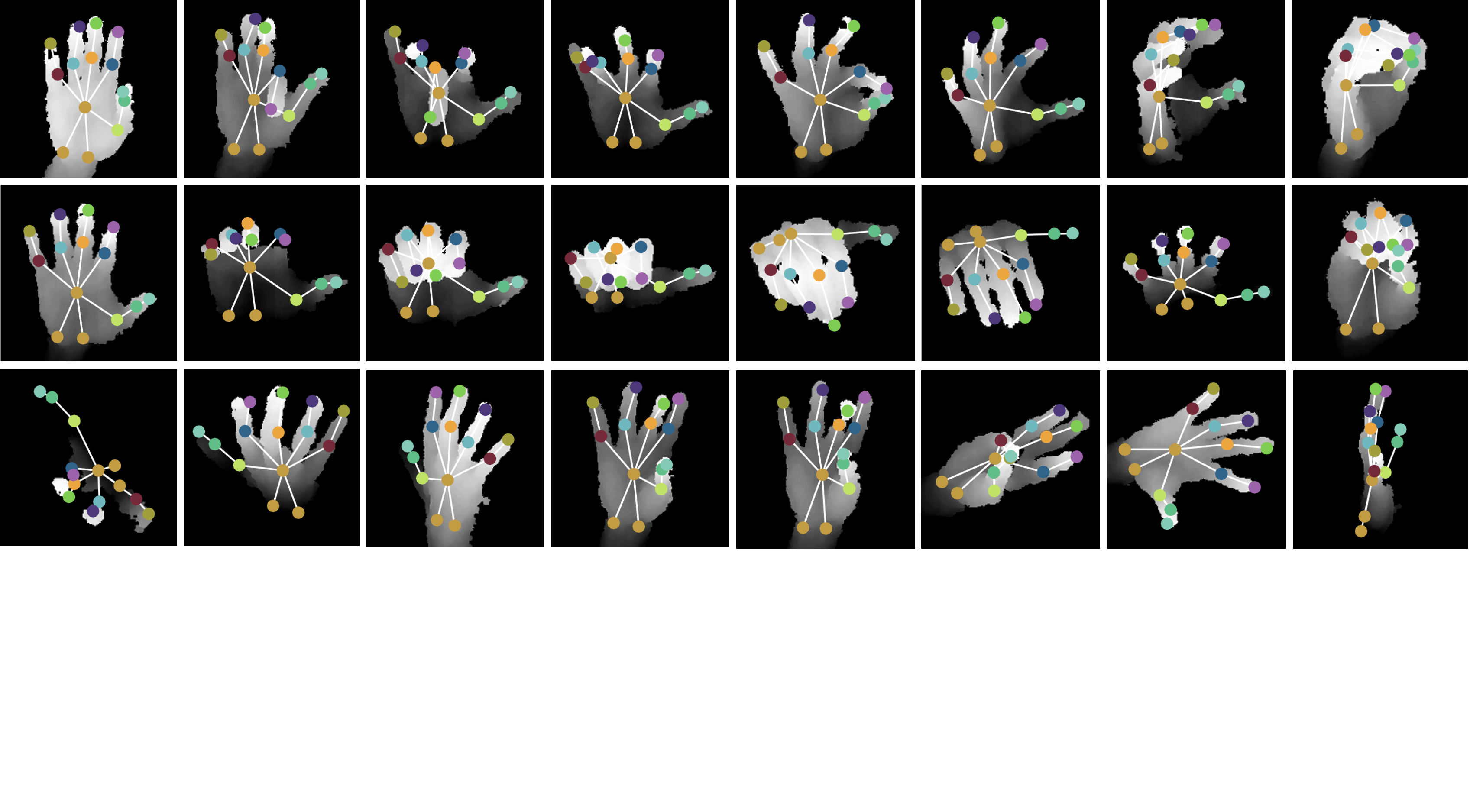}
\caption{Visualization of estimated hand skeletons produced by our network trained in a semi/weakly-supervised fashion (depth images are sampled uniformly from the NYU Hand Pose dataset).\label{fig:skeletons}}
\end{figure*}

\begin{figure*}[p!] \centering
\begin{minipage}[t]{0.6cm}
\centering
$\phantom{t}$\vspace*{-8.5cm}\\
\hfill (a)\vspace*{1.9cm}\\
\hfill (b)\vspace*{1.9cm}\\
\hfill (c)\vspace*{1.9cm}\\
\hfill (d)
\end{minipage}
\begin{minipage}[t]{17.3cm}
\centering
\includegraphics[width=0.96\linewidth]{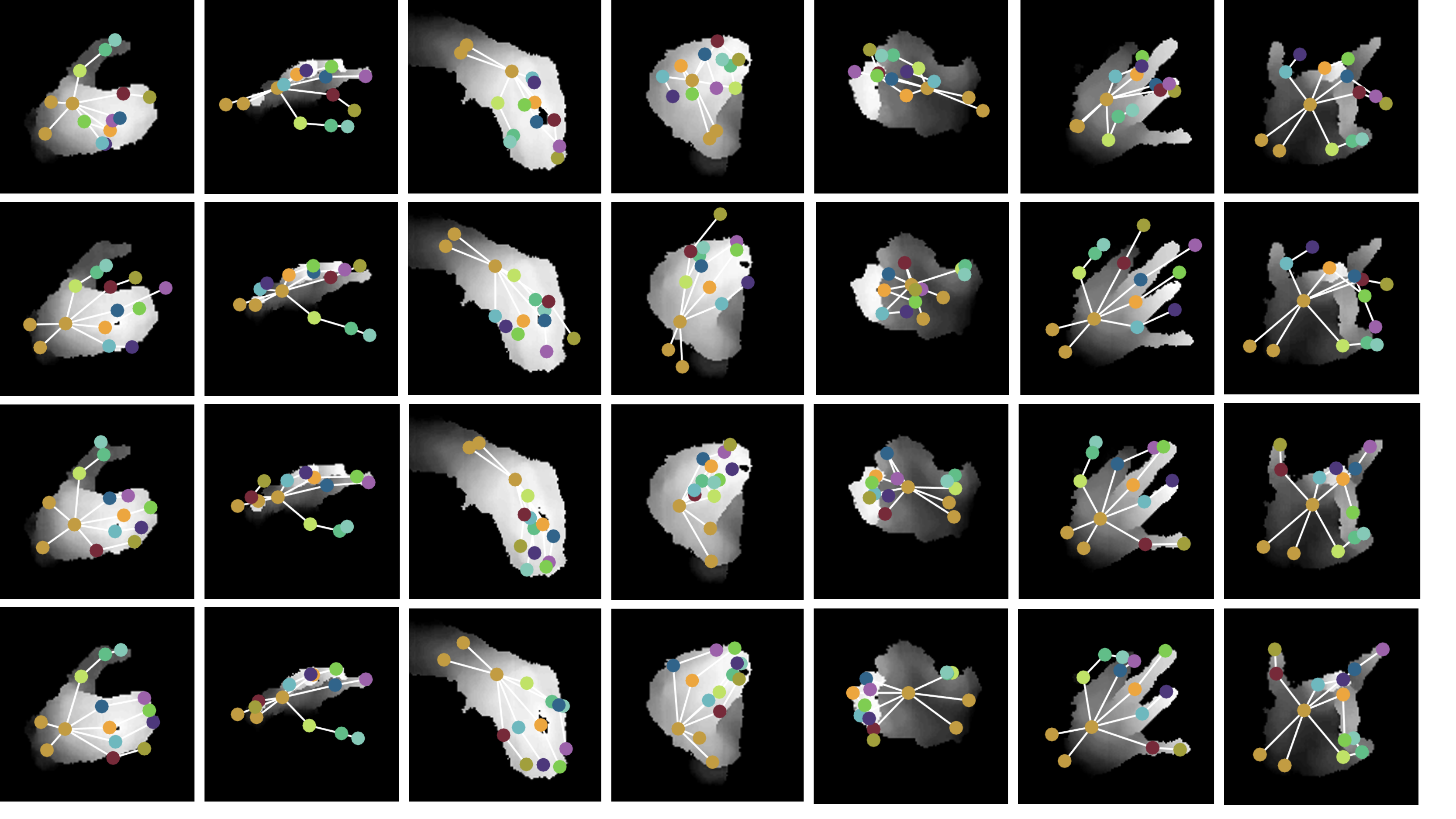}
\end{minipage}
\caption{\label{fig:vis_methods} Visual comparison of results produced on the NYU Hand Pose dataset by different 3D pose estimation methods: (a) DeepPrior~\cite{DeepPrior}, (b) Oberweger et al.~\cite{oberweger2015feedback}, (c) our method, based on the network trained in a semi/weakly-supervised fashion, (d) ground truth.}
\end{figure*}

We can see that the purely supervised baseline performs well compared to the state-of-the-art. Prior to augmenting the method with the unsupervised pipeline, we spent additional time on careful optimization of the baselines by tuning the architecture and the training regime. This appeared to be crucial, in particular the choice of batch normalization \cite{BatchNormalization2015} and Adam optimization \cite{AdamOptimization2015}. 

Figure \ref{fig:skeletons} visualizes the estimated pose skeletons corresponding to 24 input depth maps randomly sampled from the test set. Figure \ref{fig:vis_methods} illustrates performance of the proposed method on challenging examples in comparison with a number of state-of-the-art methods.
 Figure \ref{fig:plotmean} plots quantitative performance expressed as the number of frames with all joints being localized within a given distance threshold in 3D.
In these figures, we compare the proposed method with several recent state-of-the-art approaches. For the 2D hand pose estimation method proposed in \cite{TompsonSIGGRAPH2014}, we follow \cite{DeepPrior} and augment estimated $x$ and $y$ coordinates with depth values from the input depth maps. In those cases when predicted locations fall on the background, we set the corresponding z-coordinate to the median depth value of the hand.

We should note here, that the quality of network outputs can be further improved by optimization through inverse kinematics, as it has been done, for example, in \cite{TompsonSIGGRAPH2014}. However, the focus of this work is to explore the potential of pure learning approaches with no priors enforcing structure on the output.
The bottom part of the table contains non deep learning methods. In a recent work \cite{Tang2015} on optimization of hand pose estimation formulated as an inverse kinematics problem, the authors report performance similar to \cite{TompsonSIGGRAPH2014} in terms of 2D UV-error (no error in mm provided).\medskip

\subsection{Computational complexity}

\noindent
All models have been trained and tested using GPUs, except the patchwise restoration process which is  pure CPU code and not used at test time. Estimating the pose of a single hand takes 31 ms if the segmentation resolution is set to $24{\times}24$ pixels (which includes the forward passes of both networks $f_s$ (12 ms) and $f_r$ (18 ms)) and 58 ms for $48{\times}48$ segmentation outputs (40 ms for $f_s$, corresponding to the results reported in the experiments section of the paper).
Training of the segmentation network requires up to 24 hours to minimize validation error, while the regression network is trained in 20 min on a single GPU.
The results were obtained using a cluster configured with 2 x E5-2620 v2 Hex-core processors, 64 GB RAM and 3 x Nvidia GTX Titan Black cards with 6GB memory per card.


\section{Conclusion}
\label{sec:conclusion}

\noindent We presented a method for hand pose estimation based on an intermediate representation which is fused with raw depth input data. We showed that the additional structured information of this representation provides important cues for joint regression which leads to lower error. Weakly supervised learning of the mapping from depth to segmentation maps from a mixture of densely labelled synthetic data and from sparsely labelled real data is a key component of the proposed method. Weak supervision is dealt with by patch-wise alignment of real data to synthetic data performed in the space of the intermediate representation, exploiting its strong geometric and topological properties.  

\section{Acknowledgements}

\noindent This work has been partly financed through the French grant Interabot, a project of type ``Investissement's d'Avenir / Briques G\'en\'eriques du Logiciel Embarqu\'e'', and by the ANR project SoLStiCe (ANR-13-BS02-0002-01), a project of the grant program ``ANR blanc". G.~Taylor acknowledges the support of NSERC, CFI, and NVIDIA.

\section*{References}
\bibliography{references}

\end{document}